\documentclass[runningheads]{llncs}
\usepackage{graphicx, amsmath, amssymb, enumitem, float, booktabs, caption, subcaption}
\usepackage{url}
\urldef{\mailsa}\path|bkane2@ur.rochester.edu|
\urldef{\mailsb}\path|gplatono@cs.rochester.edu|
\urldef{\mailsc}\path|schubert@cs.rochester.edu|
\begin{document}
%

\title{History-Aware Question Answering in a Blocks World Dialogue System \thanks{This work was supported by DARPA grant W911NF-15-1-0542, NSF NRT Graduate Training Grant 2019-2020, and NSF EAGER Award IIS-1940981. We thank our team of volunteers for their suggestions and contributions to system evaluation.}}

%
%
\author{Benjamin Kane \and
Georgiy Platonov \and
Lenhart Schubert}
\authorrunning{B. Kane et al.}
%

\institute{University of Rochester, Rochester, NY 14627, USA \\ \mailsa\\ \mailsb\\ \mailsc}
\maketitle              
%

\begin{abstract}
It is essential for dialogue-based spatial reasoning systems to maintain memory of historical states of the world. In addition to conveying that the dialogue agent is mentally present and engaged with the task, referring to historical states may be crucial for enabling collaborative planning (e.g., for planning to return to a previous state, or diagnosing a past misstep). In this paper, we approach the problem of spatial memory in a multi-modal spoken dialogue system capable of answering questions about interaction history in a physical blocks world setting. This work builds upon a full spatial question-answering pipeline consisting of a vision system, speech input and output mediated by an animated avatar, a dialogue system that robustly interprets spatial queries, and a constraint solver that derives answers based on 3-D spatial modelling. The contributions of this work include a symbolic dialogue context registering knowledge about discourse history and changes in the world, as well as a natural language understanding module capable of interpreting free-form historical questions and querying the dialogue context to form an answer.

\end{abstract}

\section{Introduction}

AI systems have seen impressive growth in the last 10 or 20 years with respect to various specific, narrow tasks in spatial cognition and natural language processing. However, there is still a shortage of multimodal interactive systems capable of performing high-level tasks requiring understanding and reasoning. In particular, many such dialogue systems lack episodic memory -- historical recall of earlier discourse and perceived ``world" situations and events. Episodic memory is needed for supporting a sense of shared contextual awareness, and to set the stage for potential development of intelligent collaborative systems such as ones that allow diagnostic discussion of past actions, planning to re-achieve an earlier situation, repeating a past actions sequence, etc.  

Because of the complexity of most real-life tasks, the blocks world domain provides an ideal experimental setting for developing prototypes with such capabilities. Interest in the blocks world as a domain for AI research goes back as far as the 1970s, with Winograd's thesis \cite{winograd1972understanding} being one of the earliest studies, featuring a virtual environment along with text-based interaction. Winograd's system understood basic spatial relations such as one block being \textit{on} another or \textit{in the box}, and also maintained a record of block ``put-on`` actions (with their preconditions and effects), enabling it to answer questions about past actions and their purpose.

In recent years there has been a resurgence of interest in solving problems in such limited domains using modern techniques. Despite its relative simplicity, the blocks world domain motivates implementation of diverse capabilities in a virtual interactive agent aware of physical blocks on a table, including visual scene analysis, spatial reasoning, planning, learning of new concepts, dialogue management, voice interaction, and more. Recent studies in this domain have focused on learning spatial concepts in a physical blocks world setting \cite{perera2018situated,perera2018building} and applying deep learning techniques to a virtual blocks world environment \cite{bisk2018learning}. However, unlike Winograd's earlier study, these recent systems lack episodic memory and cannot reason about historical states of the world.

In this work, we extend a spatial question-answering system in a physical blocks world to allow for registering historical context and answering questions about the session history, such as \textit{``Which block did I just move?''}, \textit{``Where was the Toyota block before I moved it?''}, \textit{``Did the Target block ever touch the Texaco block?''}, \textit{``Was the Twitter block always between two red blocks?''}, \textit{``When did I last move the Starbucks block?''}, etc. Our modelling of spatial relations is based on 3-D Blender graphics representations of the objects in the blocks world; thus a straightforward approach to providing a historical record of the world would be to store successive states in this ``imagistic'' form. However, this would be intractable in terms of computation and storage in more general scenarios, where there may be object-rich scenes or indefinitely long histories. Moreover, this would appear to be cognitively implausible: Detailed visual memory of scenes in humans is quite short-lived, and a few higher-level properties suffice for humans to swiftly reconstruct more detailed representations of a scene \cite{rensink2001}. Therefore we approach this problem by maintaining a dialogue context containing symbolic knowledge about changes in the world. Used jointly with current spatial observations, this symbolic knowledge enables reconstruction of past states of the world, sufficient for answering historical questions.

\section{Related Work}

Early studies featuring the blocks world include \cite{winograd1972understanding} (as already mentioned) and \cite{fahlman1974planning}, both of which relied on a simulated environment. The latter was focused on construction planning, rather than user interaction, and as such incorporated extensive reasoning about geometric consistency and structural stability, more than descriptive aspects of the block configurations. Both of the systems maintained memory of historical states of the blocks world - the former kept a record of ``put-on'' actions, with which it could answer historical questions by reconstructing necessary knowledge, while the latter maintained a cache of known facts that it could query to improve performance. These systems demonstrated impressive planning capabilities, but their worlds were simulated, interaction was text-based, and they lacked realistic models for spatial relations.

Modern efforts in blocks worlds include work by Perera et al.~\cite{perera2018situated,perera2018building}, which is focused on learning spatial concepts (such as staircases, towers, etc.) based on verbally-conveyed structural constraints, e.g., \textit{``The height is at most 3''}, as well as explicit examples and counterexamples, given by the user. Their spatial modelling is mostly concerned with positioning of substructures such as stacks or rows with respect to each other, and sizes of rows and columns, whereas our focus is more in realistic modelling of prepositional relations, and on answering free-form questions.

In a rather different vein, the work by Bisk et al.~\cite{bisk2018learning} is concerned with learning to transduce verbal instructions, e.g., \textit{``Move McDonald's so it’s just to the right (not touching) the Twitter block"} into block displacements in a simulated environment. This system, unlike ours, relies on deep learning and does not use high-level cognitively motivated spatial relation models. The CLEVR dataset \cite{johnson2017clevr} and its modified versions, such as \cite{liu2019clevr}, lays out an explicit spatial question answering challenge that has inspired a flurry visual reasoning studies, e.g., \cite{kottur2019clevr} and \cite{mao2019neuro}, which achieves near-perfect scores on the CLEVR questions.

Common shortcomings of the deep learning approaches include a reliance on image data projected from synthetic scenes of limited variety, simplified ground-truth models of spatial relations (e.g., \textit{left} means any amount laterally to the left, regardless of depth or intervening objects, etc.), and use of domain-specific procedural formalisms for linguistic semantics. Also, while the system architectures employed could probably be carried over to other domains, there would be no carry-over of conceptual understanding or language understanding -- each new domain would require creation of new large training corpora for both the spatio-physical and linguistic aspects of the domain, with ever-increasing demands for data as the complexity of scenes being considered grows.

Finally, in relation to our focus here, we note that the recent blocks world systems do not not maintain an episodic memory or attempt reasoning about historical states of the world.

Outside of the blocks world domain, several AI systems have made use of some form of episodic memory. The TRAINS system \cite{ferguson1996} and the subsequent TRIPS system \cite{ferguson1998} were interactive dialogue-based problem solving systems in a virtual map environment. These systems maintained temporal knowledge containing facts about the planning environment, and the planning agents were able to reason about temporal aspects of plans using Allen Interval Logic \cite{allen1994}. The work reported in \cite{madl2013} implements a spatial working memory using LIDA, a symbolic cognitive architecture, in a virtual reality environment. This work represents spatial context using a combination of a grid representation of the world, and hierarchical ``place nodes'' with individual activations, which are updated based on phase changes of the grid. The performance of this system was compared to human performance on a map recall task; however, this study did not involve any ability to reason about historical spatial relations between objects. Recent deep-learning-based approaches to modelling spatial episodic memory include \cite{rothfuss2018} and \cite{franklin2019}. The former uses an unsupervised encoder-decoder model to represent episodic memory as latent embeddings, and shows that this model can allow a robot to recall previous visual episodes in a physical scene. The latter introduces a neuro-symbolic Structured Event Memory (SEM) model that is capable of segmenting events in video data and reconstructing past memory items.

\section{Task Description}

The goal of this work is to enable dialogue-based question answering about historical states of blocks arranged on a table. This includes both questions about spatially-relevant actions by the user (e.g., \textit{``Which block did I just move?''}, \textit{``What blocks did I put near the Twitter block?''}, \textit{``When did I last move the Starbucks block?''}, \textit{``What was the first block that I moved?''}, etc.), as well as questions about past spatial relations between blocks (e.g., \textit{``Where was the Toyota block before I moved it?''}, \textit{``Did the Target block ever touch the Texaco block?''}, \textit{``Was the Twitter block always on the Starbucks block?''}, etc.). Moreover, the historical module for answering these questions should be sufficiently general to be extended to more realistic domains, such as a ``room world'' containing everyday items.

This task serves two purposes. First, it motivates the augmentation of the dialogue manager with an episodic memory, and guides its integration with the overall pipeline (including dialogue management, audio-visual input/output, etc.). Experience with episodic memory design for the blocks world will allow extensions to more advanced functionalities in more general settings. Second, our overarching goal is to build a collaborative blocks world agent, capable of interactively learning structural concepts and building examples of them, relying on natural language communication with the user. Spatial episodic memory is necessary for allowing diagnostic discussion of past actions (e.g. ``You mean, next to the previous block?'', ``Where it was before, right?'', etc.), repeating past action sequences, and generally supporting a sense of shared contextual awareness.

The following example interaction between the user and the system demonstrates the kind of back-and-forth exchange our system is capable of:
\begin{itemize}[noitemsep]
    \item[] David: Would you like to ask me a spatial question?
    \item[] User: (moves a block)
    \item[] User: David, which block did I just move?
    \item[] David: You moved the Toyota block.
    \item[] User: Where was the Toyota block before?
    \item[] David: The Toyota block was between the Mercedes block and the Burger King block.
    \item[] User: Where is it now?
    \item[] David: The Toyota block is on top of the Texaco block.
    \item[] User: (moves another block)
    \item[] User: How many blocks have I moved since the beginning?
    \item[] David: You moved two blocks.
    \item[] User: Was the Twitter block always behind the Mercedes block?
    \item[] David: No.
    \item[] User: Did I move the Twitter block after I moved the Toyota block?
    \item[] David: Yes.
    \item[] User: When did I move the Toyota block?
    \item[] David: You moved the Toyota block three minutes ago.
\end{itemize}

\section{Blocks World System Overview}

The capacity for answering historical questions is built on top of an existing dialogue-based blocks world system and physical apparatus, which we describe in this section\footnote{The implementation of most of the components (apart from the Kinect blocks detector) can be found at https://github.com/gplatono/spatialQA}.

\begin{figure}[ht]
    \centering
    \begin{subfigure}{0.48\textwidth}
        \centering
        \includegraphics[height=4cm]{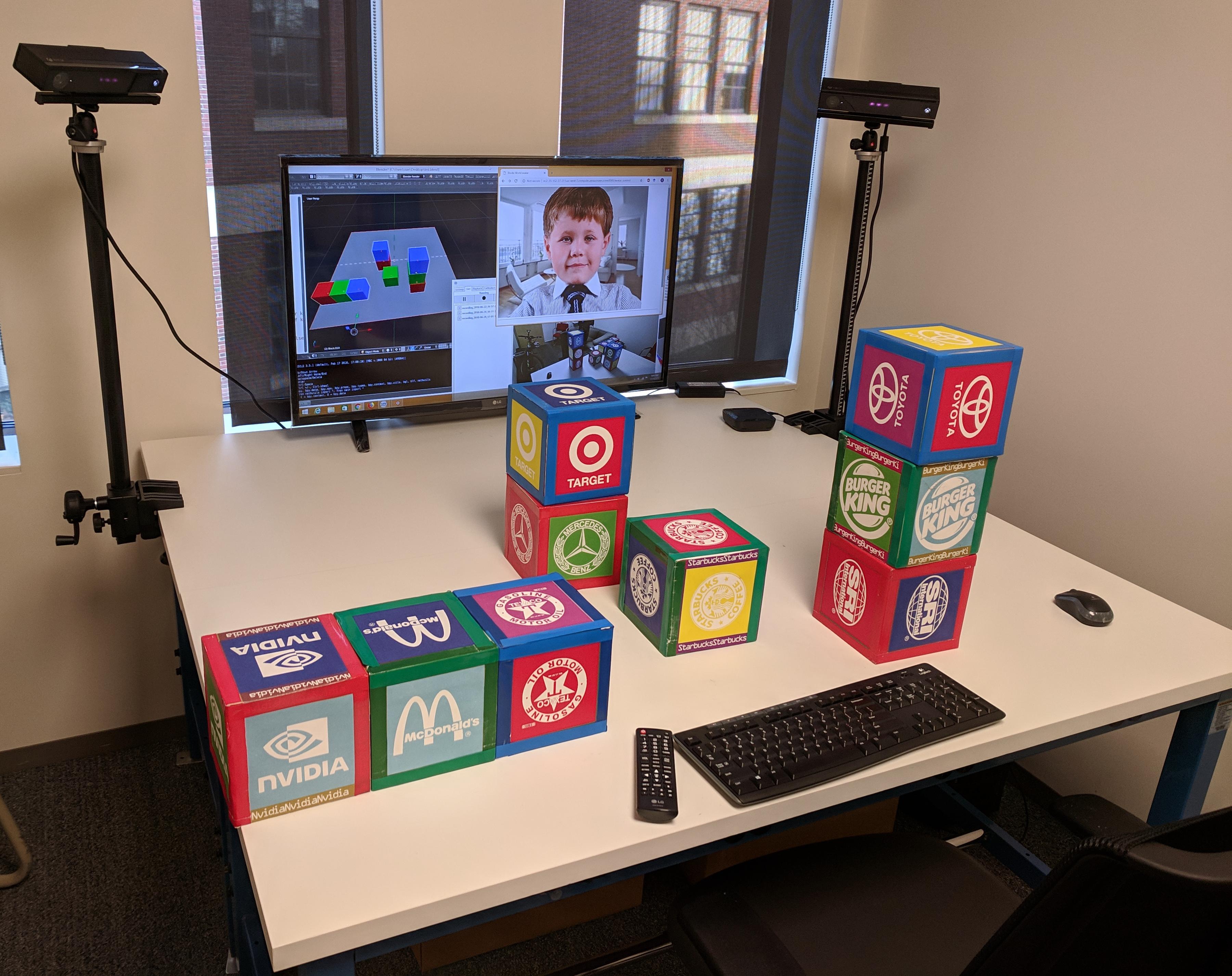}
        \caption{}
        \label{fig:bw_setup}
    \end{subfigure}
    \begin{subfigure}{0.48\textwidth}
        \centering
        \includegraphics[height=4cm]{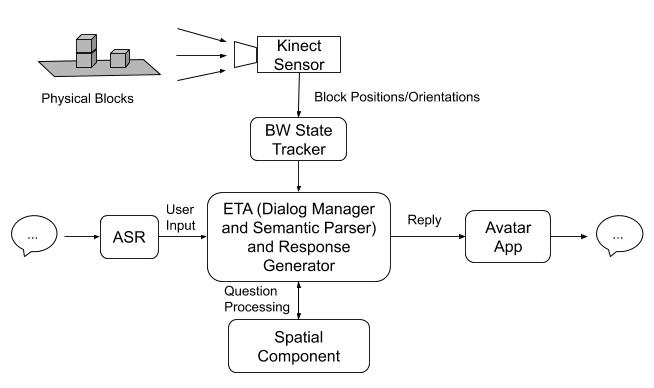}
        \caption{}
        \label{fig:bw_pipeline}
    \end{subfigure}
    \label{fig:bw}
    \caption{The blocks world apparatus setup (a), and dialogue pipeline (b). In (b), the arrows indicate the direction of interaction between the modules.}
\end{figure}

The physical apparatus (see Fig. \ref{fig:bw_setup}) is comprised of a square table surface, approximately 1.5m x 1.5m in size, several cubical blocks with 0.15m sides, two Microsoft Kinect sensors to track the state of the world, and a display for user interaction. The blocks are marked with corporate logos, such as McDonald's, Toyota, Texaco, etc., which serve as block names and allow the user and the system to uniquely identify and refer to individual blocks. The blocks are also color-coded as either red, green, or blue, using the colored stripes running along the edges of the blocks (see Figure \ref{fig:bw_setup}).

The architecture of the software component is shown in Fig.~\ref{fig:bw_pipeline}. The system uses audio-visual input and output. The block detection and tracking module periodically reads the input from the Kinect cameras and updates the block positioning information. Based on the information from the block tracking module, the physical block arrangement is modeled as a 3-D scene in Blender. All the spatial processing is performed on that model. The automatic speech recognition module, based on the Google Cloud Speech-To-Text API, is responsible for generating the transcripts of user utterances. 

For communicating back to the user, we employ an interactive avatar, David\footnote{The avatar is created using the SitePal service: https://www.sitepal.com.} It is capable of vocalizing the text and displaying facial expressions, making the flow of conversation more natural than with textual I/O.

The spatial component module together with a constraint solver is responsible for analyzing the block configuration with respect to the conditions implicit in the user's utterance. The Eta dialogue manager is responsible for {\it unscoped logical form} (ULF) generation (see subsection below) and controlling the dialogue flow and transition between phases, such as greeting, ending the session, etc. 



\subsection{The Eta Dialogue Manager and Semantic Parser} \label{sec:ETA}

The Eta dialogue manager (DM) is responsible for semantic parsing and dialogue control. Eta is designed to follow a modifiable dialogue schema, the contents of which are formulas in episodic logic \cite{schubert2000} with open variables describing successive steps (events) expected in the course of the interaction, typically speech acts by the system or the user. These are either realized directly as instantiated actions, or expanded into sub-schemas for further processing as the interaction proceeds \footnote{Intended actions obviated by earlier events may be deleted.}.

A key mechanism used in the course of instantiating schema steps, including interpretation
of user inputs, is \textit{hierarchical pattern transduction}. Transduction hierarchies
specify patterns at their nodes, with branches from a node providing alternative continuations 
as a hierarchical match proceeds. Terminal nodes provide result templates, or specify a subschema,
a subordinate transduction tree, or some other result. The patterns are simple template-like ones
that look for particular words or word features, and allow for ``match-anything", length-bounded word spans. For example, a feature-annotated word might be (spring season time-period noun name), and ``any number or words" is indicated by 0, and ``at most two words" is indicated by 2. 

As described so far, the DM resembles the dialogue manager used by the LISSA system \cite{razavi2016IVA,razavi2017ACS}. However, interpretation in that system was designed for
casual conversation, and was limited to context-dependent derivation of English \textit{gist clauses} from user inputs. The gist clauses were derived using transduction trees that take account
of prior utterances, largely eliminating context dependence in the process. This greatly
simplifies the process of finding appropriate responses to inputs -- again via transduction trees. Eta likewise uses gist clause derivation, but only for handling casual aspects of dialogue such as greetings, and for ``tidying up" some inputs in preparation for for further processing.

\begin{figure}[ht]
    \centering
    \includegraphics[width=0.9\columnwidth]{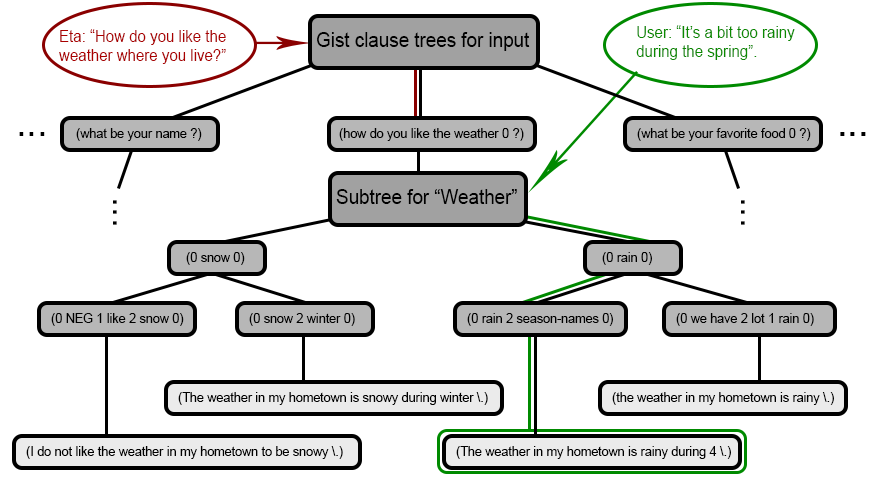}
    \caption{An example pattern transduction tree for gist clauses, with a trace for specific example inputs shown.}
    \label{fig:transduction_gist}
\end{figure}

A simplified generic example of a gist clause transduction tree is shown in Figure \ref{fig:transduction_gist}. The gist clause of Eta's previous utterance (shown in red) is used to select an appropriate subtree, which is next used to extract a gist clause from the user's utterance (shown in green). 

After extracting gist clauses, Eta also derives an \textit{unscoped logical form} (ULF) \cite{kim2019type} from the tidied-up input. ULF is closely related to the logical syntax used in schemas -- it is a preliminary form of that syntax, when mapping English to logic. ULF differs from similar semantic representations, e.g., AMR, in that it is close to the surface form of English, covers a richer set of semantic phenomena, and does so in a type-consistent way. To illustrate the approach, consider the example ``Which blocks are on two other blocks?''. The resulting ULF will be (((Which.d (plur block.n)) ((pres be.v) (on.p (two.d (other.a (plur block.n)))))) ?). As can be seen from this example, the resulting ULF retains much of the surface structure, but uses semantic typing and adds operators to indicate plurality, tense, aspect, and other linguistic phenomena. 

We extended the semantic parsing mechanism, originally aimed at deriving gist clauses, by introducing phrase-based recursion into hierarchical transduction trees. This enabled a rather novel form of compositional interpretion that is quite efficient and accurate for the domain at hand, and has proved to be readily extensible. A top-level transduction tree identifies different types of input sentences and accordingly sends them off to more specialized trees. These trees again use hierarchical pattern matching based on words and their features to identify meaningful (generally phrasal) segments of the input, such as an NP segment or a VP segment. They then dispatch the corresponding (feature-annotated) word sequences to transduction hierarchies appropriate for their phrasal types; these recursively derive and return ULF formula constituents, which are then composed into larger expressions by the ``calling" tree, and returned. At the level of individual words (or certain phrases), a lexicon and lexical routines supply word ULFs. The efficiency and accuracy of the approach lies in the fact that hierarchical pattern matching can quite accurately segment utterances into meaningful parts, often relying on automatically added syntactic and semantic features, so that the need for recursive backtracking rarely arises. Also some transductions may remove or ignore extraneous words (such as the first two words of ``OK, so, when did the ...''), improving robustness.

\begin{figure}[ht]
    \centering
    \includegraphics[width=\columnwidth]{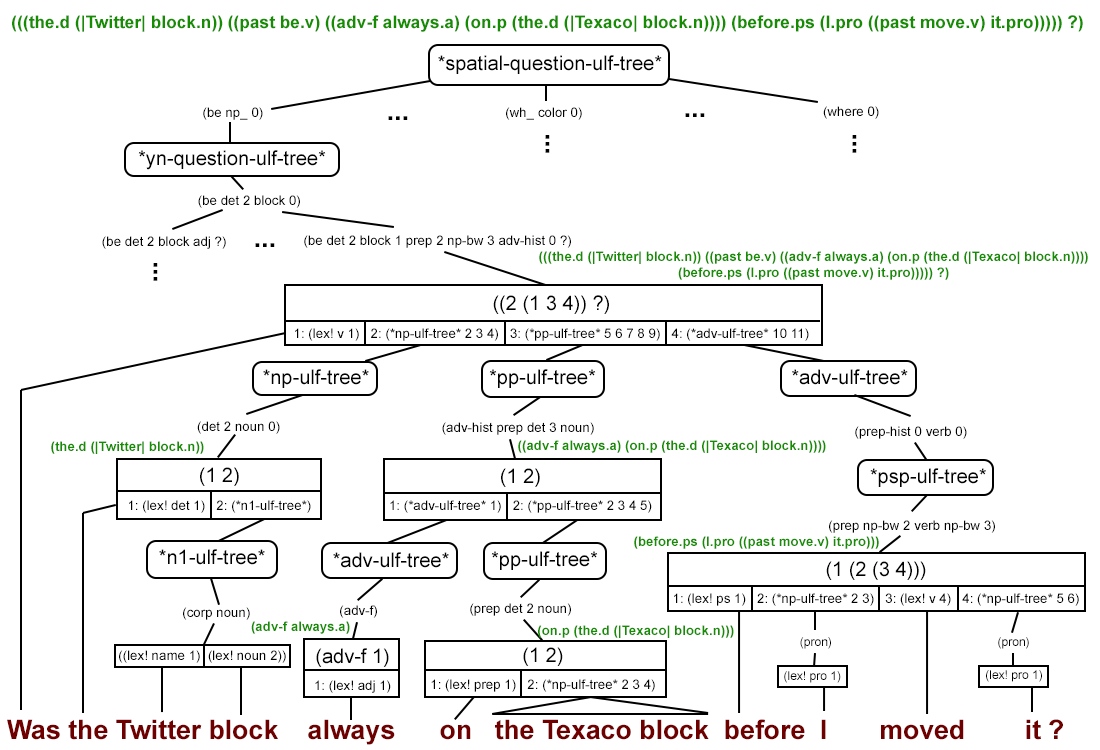}
    \caption{An example ULF parse, with the input shown in red, and the resulting ULF (at each composition step) shown in green. The nodes with rectangles represent ULF composition nodes, where the numbers in the upper box correspond to the indices of the lower boxes (if there is no upper box, the constituent ULFs are simply concatenated). All other nodes are patterns to be matched to the corresponding span of input text.}
    \label{fig:historical_ulf_parse}
\end{figure}

An example of a transduction tree being used for parsing a historical question into ULF is shown and described in Figure \ref{fig:historical_ulf_parse}. As in the example mentioned above, the resulting ULF retains much of the surface structure, but uses semantic typing and adds operators to indicate plurality, tense, aspect, and other linguistic phenomena. Additional regularization is done with a limited coreference module, which can resolve anaphora and referring expressions such as ``it'', ``that block'', etc., by detecting and storing discourse entities in context and employing recency and syntactic salience heuristics. 

\section{Historical Question-Answering}

To answer historical questions, the blocks world agent requires two related functionalities: First, the DM must maintain a dialogue context, including (besides basic indexical knowledge such as current time, location, and dialogue participants) the discourse history, a list of past referents for reference resolution, and spatial episodic memory. Secondly, the DM must be able to robustly parse historical questions into the logical form described above, and consequently resolve the resulting semantic interpretation into operations over the episodic memory.

\subsection{Dialogue Context}\label{sec:context}

As noted earlier, the spatial component of an interaction memory might consist of detailed visual or vector-based memory representations that the system can query, or else it might be implemented as a high-level symbolic memory enabling reconstruction of past scenes. Inspired by Winograd's early work in \cite{winograd1972understanding}, and mindful of the cognitive considerations already cited \cite{rensink2001}, we chose to adopt the latter approach. 

As the spatial question answering session progresses, the vision system records the centroid coordinates of blocks and block moves in real time, thresholded to avoid registering noise as block moves. On the DM side, a ``perceive-world'' action in the schema causes the DM to request perceptions (represented in ULF) from the vision system. These perceptions currently consist of block location propositions of the form ($|$Twitter$|$ at-loc.p (\$ loc ?x ?y ?z)),\footnote{``$|$Twitter$|$'' exemplifies a block name.} and block move propositions of the form ($|$Twitter$|$ ((past move.v) (from.p-arg (\$ loc ?x1 ?y1 ?z1)) (to.p-arg (\$ loc ?x2 ?y2 ?z2)))). In principle our formalism also allows named locations, e.g., ($|$Twitter$|$ at-loc.p $|$Loc1$|$), though this is not yet implemented.

We rely on a simple linear, discrete time representation (possible future modifications are discussed in Section \ref{sec:discussion}). The DM stores a symbol denoting the current time, with $|$Now0$|$ representing the time at which the dialogue is initialized. Each sequential action in the world causes the DM to ``update'' its time twice corresponding to the time during which the move is in-progress and the time at which the move has finished. That is, if the DM denoted the initial time with $|$Now0$|$, a block move would cause it to update its time to $|$Now1$|$ (the in-progress time), and then to $|$Now2$|$ once the move has finished. These temporal symbols are related to each other via propositions of the form ($|$Now1$|$ before.p $|$Now2$|$) and ($|$Now2$|$ after.p $|$Now1$|$) stored in the context.\footnote{Record structures specifying current world time are also attributed to these symbols; these are used in forming answers to ``when'' questions.} The fact (($|$Twitter$|$ ((past move.v) (from.p-arg (\$ loc ?x1 ?y1 ?z1)) (to.p-arg (\$ loc ?x2 ?y2 ?z2)))) * $|$Now1$|$) is stored in the dialogue context, where `*' is the episodic ``true in'' operator described in \cite{schubert2000}. User utterance actions are similarly stored in the context.

Based on this context, the DM can efficiently reconstruct a scene at any past time by backtracking from currently observed block locations, as well as use these reconstructed scenes to evaluate spatial relationships between blocks in a "``rough-and-ready'' way, i.e., using approximate calculations of spatial relations based on block centroid coordinates, as opposed to the detailed spatial computations supported by the visual blocks world system.

\subsection{Interpreting Historical Questions}\label{sec:hqinterpret}

Following a successful parse of a historical question by the semantic parser described previously, historical modifiers in a ULF will be indicated by constituents of type ``adv-e'' (event adverbial, e.g., (adv-e (during.p (the.d move.n)))), ``adv-f'' (frequency adverbial, e.g., (adv-f (three.a (plur time.n)))), or ``adv-s'' (sentence adverbial, e.g., (adv-s (after.ps ($|$Twitter$|$ (past move.v))))).


The algorithm the DM uses to answer historical questions is as follows: starting from the present time, the algorithm iterates over past times, reconstructing the scene at each one using stored knowledge about moves. At each time, the algorithm computes and stores a list of salient facts (i.e. propositions about spatial relations or actions which held at that time) depending on the subject, object, predicate, question category, and polarity of the query sentence. Furthermore, temporal constraints are applied to filter these times (in the manner described below) to obtain a final list of times with relevant attached facts.

\begin{figure}[ht!]
    \centering
    \includegraphics[width=\columnwidth]{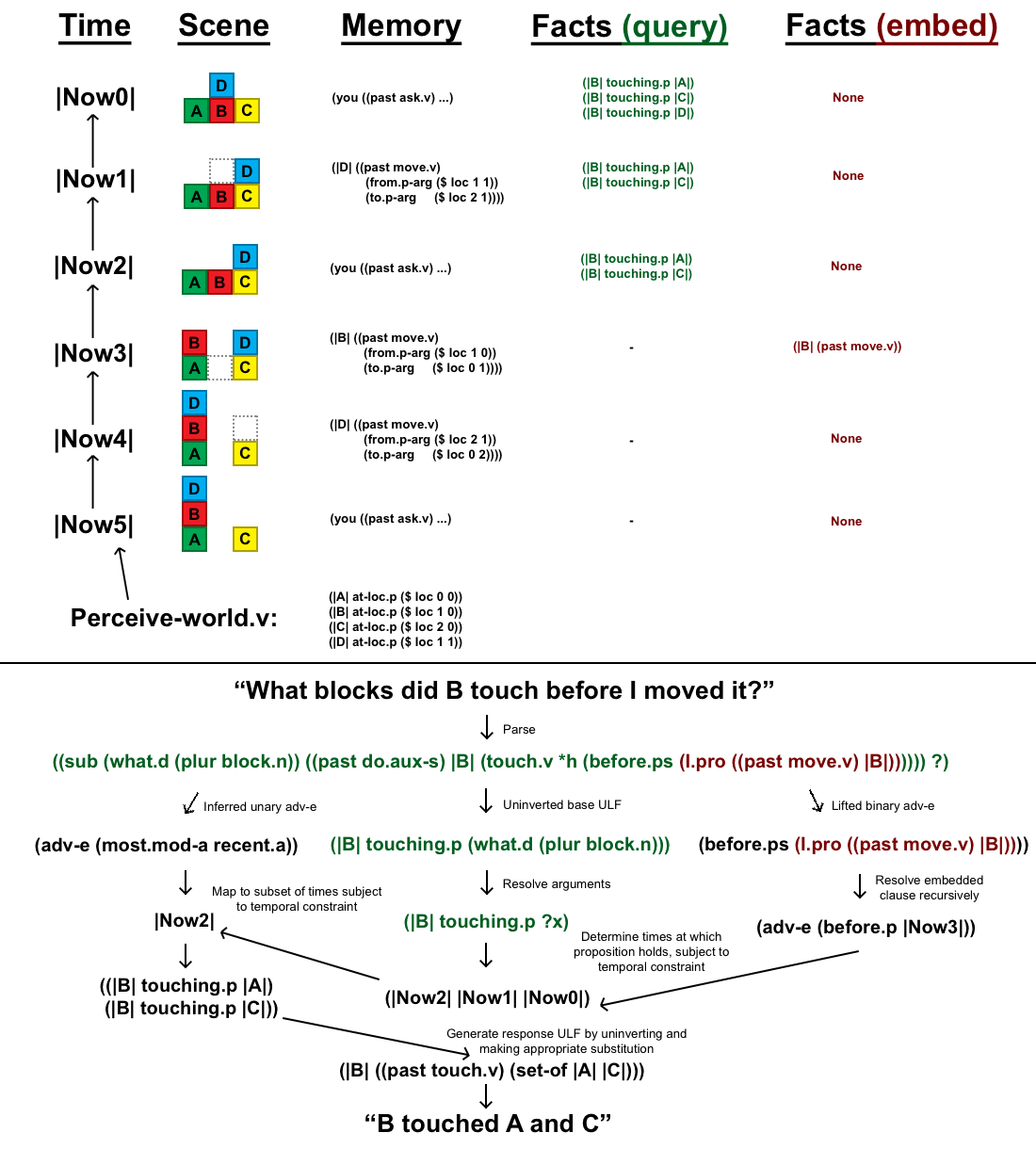}
    \caption{A simplified example of how the context is represented and how the DM uses the context to compute relations given temporal constraints (top half), and an example of the DM determining an answer from a specific historical query (bottom half). Note that, although the visual scenes are shown for reference, the DM does not actually store detailed visual memories; it only stores the symbolic facts in the ``Memory'' column.}
    \label{fig:historical_qa}
\end{figure}

The semantic types of these temporal and frequency modifiers allow them to be lifted to the sentence level \cite{kim-schubert-2017-intension}. Temporal constraints expressed by modifiers may be binary, e.g., (adv-e (before.p $|$Now4$|$)), or unary, e.g. (adv-e recent.a). A binary constraint takes a temporal entity as an argument and maps it to a truth value, depending on whether the given relation holds with the object of the constraint. This is used by the algorithm described above to filter out each time at which the binary constraint does not evaluate to true. However, first the binary constraint needs to resolve its object constituent (which could be a simple noun phrase or an embedded clause). This is done using a recursive call of the algorithm described above, which maps the object ULF to a list of times, treating any modifiers in the noun phrase or embedded clause as temporal constraints.

A unary constraint takes a set of times and maps it to a subset (possibly null) of these times. For example, the ``recent'' constraint above picks out the subset of times that are within some fixed threshold to the present time. Frequency constraints such as twice.adv-f or (adv-f (three.a (plur time.n))) are similar to unary constraints in that they take a set of times and return a subset of these, though their behavior is slightly more complicated -- they pick out all times for which the salient facts attached to that time are also attached to at least N unique times, inclusive. For (adv-f always.a), N is taken to be the size of the set of times being filtered (so that only facts that are attached to every time in the set are obtained).

Each constraint may also be modified by a ``mod-a'' modifier, e.g., (adv-e (just.mod-a recent.a)), which modifies how that constraint is applied. In the case of ``just recently'', the singular most recent time is picked out.

Historical questions don't necessarily involve sentence-level adverbial modifiers, as the temporal content could be embedded within a noun phrase, as in ``What was the first block that I moved?''. In this case, the DM will resolve this reference to a particular block by calling the above algorithm recursively, treating the noun pre- and post-modifiers as temporal constraints, and using the facts attached to the resulting times.

Once a list of final times and the corresponding facts/relations have been obtained, an answer is generated by making the appropriate substitutions in the query ULF (e.g. a wh-pronoun for the subject of a relation), applying syntactic transformations (e.g., uninverting questions and removing auxiliary verbs such as ``do''), and converting this to surface form.

The DM additionally has a limited module for generating and responding to pragmatic inferences, based on the work in \cite{kim-etal-2019-generating}. In the case where either no times or no relations are obtained, and the question contains a presupposition (e.g. ``What block was the Twitter block on?'' carries the presupposition that the Twitter block was on some block), the DM will attempt to respond by negating the inferred presupposition (e.g. ``The Twitter block wasn't on any block.'').

A full example of answering a historical question (using a simplified scene) is shown in Figure \ref{fig:historical_qa}. The extraction of answer relations, given the query ULF and the generation of the answer ULF, are shown in the bottom half of the figure, while the scene reconstruction and computation of relevant facts/relations is depicted in the top half of the figure.

Note that the example in Figure \ref{fig:historical_qa} is actually fairly ambiguous; the answer could be ``A, D, C'' or ``A, C'' depending on whether someone reads the query as meaning any blocks that B \textit{ever} touched before it was moved, or only the blocks that it touched directly before the move. In fact, we found that many natural historical questions are similarly under-specified, presenting a major source of difficulty. To deal with this issue, the DM's pragmatic module attempts to infer temporal constraints in these ambiguous cases -- in this particular example, Eta would infer the constraint ``most recently'', unless the user explicitly specifies otherwise in their query.

\section{Evaluation and Discussion}\label{sec:discussion}
Though our work is grounded in a physical blocks world system, the COVID-19 pandemic made an on-site user study impossible. Therefore we resorted to developing a virtualized environment that mirrors our setup, and used it to collect the evaluation data. Only the physical blocks tracker and the audio I/O were disabled in the modified system. All the crucial components evaluated in this work, namely, the parser, the dialog manager, and the historical question-answering subsystem based on the world state memory were not changed, so the results of the user study are not invalidated.

We enlisted the help of 4 student volunteers to test the capabilities of the system, including both native and non-native English speakers. The participants were instructed to move the blocks around and ask general questions about relationships and changes in the world; no restrictions on wording were imposed. After the system displayed its answer, the participants were asked to provide feedback on the quality of the answer, by marking the system's answer as correct, partially correct or incorrect. Each participant contributed at least 100 questions.

Each session started with the blocks positioned in a row at the front of the table. The participants were instructed to reposition or stack up the blocks arbitrarily in the course of the question-answer session, to test the robustness and consistency of the spatial models. The data is presented in Table \ref{evaltab}. A few malformed questions were excluded when computing accuracy.

\begin{table}[ht]
\caption{Evaluation data.}
\centering
\begin{tabular}{|c|c|}
    \hline
    Total number of questions asked & 496 \\
    \hline
    Well-formed questions & 472 \\
    \hline
    Correct answers & 250 (53\% of 472) \\
    \hline
    Partially correct answers & 25 (5\% of 472) \\
    \hline 
    Incorrect answers & 197 (42\% of 472)\\
    \hline 
    Number of correctly parsed questions & 445 (94\% of 472) \\
    \hline
    Accuracy (correct + partially correct) & 58\%\\
    \hline
\end{tabular}
\label{evaltab}
\end{table}

We find these preliminary results encouraging, given the complexity of the task and the unrestricted form of the questions, though there is still much room for future improvement. A little above half of Eta's answers were judged to be fully correct, with accuracy rising to 58\% when including partially correct answers. We find that the semantic parser itself is very reliable, with 94\% of grammatical sentences being parsed correctly.

We observe that historical questions are, in general, far more pragmatically loaded than simple spatial questions, and judgements involve high degrees of subjectivity. A major source of error is in the handling of under-specified historical questions, as described in Section \ref{sec:hqinterpret}. There are many nuances to how humans naturally interpret these, that are difficult to consistently capture with simple pragmatic rules. For example, Eta will plausibly interpret ``What blocks did I move before the Twitter block?'' as meaning ``What blocks did I move \textit{shortly} before I moved the Twitter block?'' (especially if the move of the Twitter block was very recent); however if the user instead asks ``How many blocks did I move before the Twitter block?'', it seems that the questioner probably means ``How many blocks did I \textit{ever} move before I moved the Twitter block?''. Currently, Eta would add ``recently'' for the latter case, which would be incorrect. 


\section{Future Work}

In future work, we aim to investigate the pragmatic phenomena discussed in Section \ref{sec:discussion} in more detail and to improve the pragmatic inference module to handle these cases correctly, as well as carrying out more detailed analyses of other sources of error.

In addition, as questions in the blocks world domain tend to exhibit a fairly simple tense structure, we encountered issues with some of the more complex questions as a result of our simplifying assumption of discrete linear time. In future work, we plan to look into the use of more general temporal reasoning systems such as the tense trees described in \cite{schuberttense1994} to enable the system to handle different aspects and more complex embedded clauses more robustly.

Finally, as described in Section \ref{sec:context}, our system approximates each object in memory in terms of the position of its centroid with a cubical bounding box around it. Although this approach is justifiable in view of people's generally rather vague and unreliable recall of spatial relations, it can also lead to deviations from human judgments, especially for configurations 
conceptualized by people in terms of larger shapes. For example, if multiple blocks are arranged in a crescent shape, where that crescent surrounds an additional block nearly but not quite in contact with it, a person would remember that the interior block was not actually touching the nearest block of the crescent, whereas our centoid-based computation might well decide that they were touching.
In our continuing work on natural language interaction in the blocks world to allow for teaching and learning larger-scale structural concepts, and also generalizing to a more realistic ``room world'' (see \cite{platonov2018computational}), we are developing a set of object schemas for objects in the domain, using much the same formalism as for the dialogue schemas described above (but augmented with 3-D prototypes). Dealing with historical questions in such settings will require enrichment of episodic memory representations and of the linguistic and spatial reasoning mechanisms for interacting intelligently with a user. 

\section{Conclusion}
We have augmented a spatial question answering system in a physical blocks world system with the ability to answer free-form historical questions using a symbolic dialogue context, keeping track of a record of block moves and other actions. A pattern-driven, compositional semantic parser allows historical questions to be parsed into a logical form, which is then used in conjunction with the historical context model to derive and generate answers. We obtained an accuracy of 58\%, which we believe is a reasonable preliminary result given the free-form and often under-specified nature of the historical questions that users asked, though it also leaves much room for improvement. Overall, the pragmatic richness and complexity that we've observed in historical question-answering suggests that further work in this under-studied area is likely to be fruitful.


\bibliographystyle{splncs04}
\bibliography{main}
\end{document}